\documentclass{article}
\usepackage{kr}

\usepackage{times}
\usepackage{soul}
\usepackage{url}
\usepackage[utf8]{inputenc}
\usepackage[small]{caption}
\usepackage{graphicx}
\usepackage{amsmath}
\usepackage{amsthm}
\usepackage{amssymb}
\usepackage{booktabs}
\usepackage{xspace}
\usepackage{enumerate}
\usepackage{tikz}
\usepackage{natbib}
\hypersetup{nolinks=true}

\usepackage[noend,linesnumbered,ruled]{algorithm2e}
\usepackage{booktabs}
\usepackage{amsmath}
\usepackage{amsfonts}
\usepackage{dsfont}
\usepackage{bbold}

\newcommand{\reals}{\mathbb{R}}

\newcommand{\Omit}[1]{}

\newcommand{\multiset}[1]{\{\!\!\{ #1 \}\!\!\}}

\newcommand{\agg}{\emph{agg}\xspace}
 \newcommand{\comb}{\emph{comb}\xspace}

\newcommand{\C}{\mathsf{C}}

\newcommand{\CC}{\ensuremath{\C_2}\xspace}
\newcommand{\CCC}{\ensuremath{\C_3}\xspace}

\newcommand{\mlp}{\mathbf{MLP}\xspace}

\newcommand{\tup}[1]{\langle #1 \rangle}

\newcommand{\mminus}{\hspace{-.05em}\raisebox{.15ex}{\footnotesize$\downarrow$}}

\newcommand{\GT}[1]{#1{\,>\,}0}

\newcommand{\DEC}[1]{#1\mminus}

\newcommand{\prule}[2]{#1  \mapsto #2}

\title{Learning Generalized Policies Without Supervision Using GNNs}

\author{%
  Simon St\r{a}hlberg$^1$\and
  Blai Bonet$^2$\and
  Hector Geffner$^{3,2,1}$ \\
\affiliations
  $^1$Link\"{o}ping University, Sweden\\
  $^2$Universitat Pompeu Fabra, Spain\\
  $^3$Instituci\'o Catalana de Recerca i Estudis Avan\c{c}ats (ICREA), Barcelona, Spain\\
\emails
  simon.stahlberg@liu.se,
  bonetblai@gmail.com,
  hector.geffner@upf.edu
}

\bibpunct{(}{)}{;}{a}{,}{,}

\begin{document}

\maketitle

\begin{abstract}
We consider the problem of learning generalized policies for classical planning domains
using graph neural networks from small instances represented in lifted STRIPS.
The problem has been considered before but the proposed  neural architectures are complex and the results are often mixed.
In this work, we use a simple and general GNN architecture and aim at obtaining
crisp experimental results and a deeper understanding: either the policy greedy in the learned value function
achieves close to 100\% generalization over  instances larger  than those used  in training, or the failure must be
understood, and possibly fixed,  logically. For this, we exploit the relation established
between the expressive power of GNNs and the \CC fragment of first-order logic (namely, FOL
with 2 variables and counting quantifiers). We find for example that domains with general policies
that require  more expressive features can be solved with GNNs once the states are extended
with suitable "derived atoms" encoding role compositions  and transitive closures that
do not fit into \CC. The work follows the GNN approach for learning optimal general policies
in a supervised fashion \citep*{simon:arxiv2021}; but  the learned policies are no longer required
to be optimal (which expands the scope,  as many planning  domains do not
have general optimal policies) and are learned without supervision. Interestingly,
value-based reinforcement learning methods that aim to produce optimal policies,
do not always  yield policies that  generalize, as the goals of optimality and
generality are in conflict in domains where optimal planning is NP-hard.
\end{abstract}

\section{Introduction}

Generalized planning is concerned with the computation of  general policies  for  families of planning instances over the same domain
that  span different state spaces. For example, a general policy for solving Blocks problems
can place all blocks on the table and  stack then the desired  towers, bottom up, one at at time.
The formulation and the computation of general policies is particularly
interesting at it involves ideas from planning, knowledge representation, and learning. Indeed,
the language for representing the
general policies is key, in particular in  domains where the set of ground actions change from instance
to instance \citep{bonet:ijcai2018}. Also   learning  policies from examples
has been found to be simpler than synthesizing them from  specifications
\citep{khardon:generalized,srivastava08learning,bonet:icaps2009,hu:generalized,BelleL16,anders:generalized}.
In planning, it is common to approach the problem assuming that  domain predicates are  known,
while some  deep learning and deep reinforcement learning approaches  address  the problem with no domain knowledge,
representing the states, for example,  as 2D images \citep{babyAI,minigrid:amigo,procgen}.

In this paper, we consider the problem of learning generalized policies for classical planning domains
using  \emph{graph neural networks}  \citep{gori:gnn,book:gnn}
from small instances represented in lifted STRIPS.  The  problem has been considered before  but
using    neural architectures that are more  complex and with results that are often less crisp,
involving in certain cases  heuristic information or search
\citep{sylvie:asnet,mausam:dl,karpas:generalized,sid:generalized,trevizan:search}.
We use a simple and general GNN architecture and aim at  obtaining  crisp experimental results
and a deeper  understanding: either the policy greedy  in the learned value function achieves close to
100\% generalization over  instances larger than those used  in training, or the failure must be
understood and,  possibly fixed,  using logical methods. For this, we exploit the relation
between the expressive power of GNNs and   the two-variable  fragment of first-order logic with counting,
\CC,  that includes the standard description logics  \citep{barcelo:gnn,grohe:gnn}.
Description logic features have been used indeed for expressing  general policies and general value functions
\citep{martin:kr2000,fern:generalized,bonet:aaai2019,frances:ijcai2019,frances:aaai2021}.
We find for example that domains with general policies that require more expressive features
can be solved with GNNs once the states are extended with suitable "derived atoms"
for encoding role compositions and transitive closures that do  not fit into \CC.

The work follows  the GNN approach for learning optimal general policies in a supervised fashion
\citep{simon:arxiv2021} but the learned policies are no longer required to be
optimal, which expands the scope of the approach, as many planning  domains do not admit  general optimal policies,
and  are learned without supervision. The learning problem becomes the problem of learning a value function $V$
that can be applied to the states $s$ of any domain instance, such that the greedy policy in $V$ solves the training instances.
Versions of this idea have been used  in combinatorial settings \citep{frances:ijcai2019,frances:aaai2021}.
Interestingly, value-based reinforcement learning methods that aim to  produce  optimal value functions $V=V^*$
are shown not to  generalize as well in domains that admit (non-optimal) general policies
but where optimal planning is NP-hard. 

The rest of the paper is organized as follows. First we discuss related research,
then cover the background (classical planning, general policies and value functions, and GNNs)
and the actual GNN architecture and loss functions used for learning.
This is followed by the experimental section, analyses, and a summary.

\section{Related Work}

Some related research threads  are the following.

\subsubsection*{Generalized planning (GP).} Formulations of generalized
planning differ in the way in which  general policies are represented;
most often, as logic programs, finite-state controllers, or  programs with
loops  \citep{khardon:generalized,srivastava08learning,bonet:icaps2009,hu:generalized,BelleL16,anders:generalized}.
In all cases, the most compact policies that manage to solve a family of examples are sought,
and the  key question is how the space of possible programs or controllers is defined.

\subsubsection*{GP with logical features.} An alternative approach is to define the general policies
as collection of rules over a set of logical features \citep{bonet:ijcai2018}, often
derived  from the domain predicates using a  description logic grammar \citep{martin:kr2000,fern:generalized}.
Recent methods learn such  policies   from  pools  of such features \citep{bonet:aaai2019,frances:aaai2021};
in some cases, by learning  value functions    \citep{frances:ijcai2019}.
The  Boolean and  numerical features are closely related to the variables
used  in qualitative numerical planning models   \citep{sid:aaai2011,bonet:qnps}.

\subsubsection*{Generalized  policies using deep learning.} Deep learning and deep reinforcement learning methods
have been used to compute general policies from sampled problems without having to predefine the space
of possible features. In some cases, the  planning representation of the domains is used  \citep{sylvie:asnet,mausam:dl,karpas:generalized};
in other cases, it is not \citep{sid:sokoban,babyAI,minigrid:amigo,procgen}.  Also  in some cases, the learning is supervised;
in others, it is based on reinforcement learning \citep{bertsekas:dp,sutton:book,drl-book}.
The neural networks learn to map states into a feature representation that is mapped into the  value or  policy associated to the  state.

\subsubsection*{GNNs and logic.} A graph neural network learns to map vertices of a graph
into feature representations that can be aggregated and fed into a feedforward neural network
for classifying graphs, and more generally, for computing functions over graphs independently of their
size \citep{gori:gnn,book:gnn}. Since the computational model is based on message passing, GNNs cannot distinguish all pairs of graphs that are not isomorphic
but can distinguish those that are distinguished by the WL coloring procedure \citep{morris:gnn-wl,jegelka:gnn-wl}.
These correspond in turn to those that can be distinguished by formulas in the two-variable fragment of first-order logic with counting quantifiers,
\CC, which includes the standard description logics \citep{barcelo:gnn,grohe:gnn}.

\subsubsection*{GNNs and optimal general policies.}
\citet*{simon:arxiv2021} use GNNs to learn optimal general policies in a supervised fashion from
targets $V^*(s)$ and  sampled states $s$, taking advantange of a GNN architecture
introduced for learning to solve Max-CSPs \citep{grohe:max-csp}, extended to the
more general relational structures underlying planning states where objects  define
the universe, predicates  define the relations, and atoms define their denotations.
In this work, we build on these results to learn general policies that are not necessarily optimal
(and which hence cover more domains) without supervision and without having to predefine a pool of
features \citep{frances:ijcai2019}.

\section{Classical Planning}

A classical planning problem  is a pair  $P\,{=}\,\tup{D,I}$ where
$D$ is a  first-order  \emph{domain} and $I$ contains information about the instance \citep{geffner:book,ghallab:book,pddl:book}.
The domain $D$ contains a set of predicate symbols $p$ and a set of  action schemas with preconditions
and effects given by atoms $p(x_1, \ldots, x_k)$ where each $x_i$ is an argument of the schema.
An instance is a tuple $I\,{=}\,\tup{O, \textit{Init},\textit{Goal}}$ where $O$ is a set of object
names $c_i$, and $\textit{Init}$ and $\textit{Goal}$ are  sets  of \emph{ground atoms} $p(c_1, \ldots, c_k)$.

A classical problem $P\,{=}\,\tup{D,I}$ encodes  a  state model $S(P)=\tup{S,s_0,S_G,\textit{Act},A,f}$
in compact form where the states $s \in S$ are sets of ground atoms from $P$,
$s_0$ is the initial state $I$, $S_G$ is the set of goal  states $s$ such that $S_G \subseteq s$,
$\textit{Act}$ is the set of ground actions in $P$, $A(s)$ is the set of ground actions
whose preconditions are (true) in $s$, and $f$ is the transition function
so that $f(a,s)$ for $a \in A(s)$ represents the state $s'$ that follows
action $a$ in the state $s$.   An action sequence $a_0, \ldots, a_{n}$
is applicable in $P$ if $a_i \in A(s_i)$ and  $s_{i+1}=f(a_i,s_i)$, for
$i=1, \ldots, n$, and it is a plan if $s_{n+1} \in S_G$.
The \emph{cost}  of  a plan is   assumed  to be given by its length
and  a plan is \emph{optimal} if there is no shorter plan.

The  representation of  planning problems $P$ in two parts $D$ and $I$, one that is general,
and the other that is specific, is essential for defining and computing general
policies, as  the instances are assumed to come all from the same domain.
Recent work has addressed the problem of learning the action schemas and predicates
\citep{locm,asai:fol,bonet:ecai2020,ivan:kr2021}.

\section{General Policies and Value Functions}

One approach for expressing general policies is as rules $C \mapsto E$ where the condition $C$ and the effect $E$ are defined in terms of
state features \citep{bonet:ijcai2018}. State features or simply,  features,  refer to functions $\phi$ over the state,
and Boolean and numerical features refer to state functions that return Boolean and numerical values.
For example, a    general policy for  clearing a block $x$ can be expressed in terms of  the two  features $\Phi=\{H,n\}$,
where $H$ is a true in a state if a block is being held,  and $n$ represents the number of blocks above
$x$. The   policy rules are
\begin{alignat}{1}
  \label{eq:generalP}
  \prule{\neg H, \GT{n}}{H, \DEC{n}} \quad\hbox{ , }\quad \prule{H}{\neg H}
\end{alignat}
that say that,   when the gripper is empty and there are blocks above $x$, any action that decreases $n$ and
makes $H$ true should be selected, and that  when the gripper is not empty,
any action that  makes $H$ false and does not affect $n$ should be selected.
General policies of this  form can  be learned without supervision
by solving a combinatorial optimization problem $T({\cal S},{\cal F}$)
where ${\cal S}$ is a set of sampled state transitions  and $\cal F$ is a  large but finite pool of
description logic features obtained from the domain predicates \citep{bonet:aaai2019,frances:aaai2021}.

Another  way to represent (general) policies is by means of (general) value functions.
In  dynamic programming and RL \citep{bellman:dp,sutton:book,bertsekas:dp},
a value function $V$ defines  a (non-deterministic) \emph{greedy policy} $\pi_V$ that selects in a state $s$
any possible successor state $s'$ with minimum $V(s')$ value under the assumption that actions are deterministic and have
the same cost.  A  policy $\pi$ solves an instance $P$ if the state transitions compatible with $\pi$,  starting with the initial state,
eventually  end up in a  goal state.
If $V$ is optimal, i.e., $V=V^*$,  the greedy policy $\pi_V$ is optimal too, selecting state transitions along optimal paths.

General value functions for a class of problems are   defined in terms of features $\phi_i$ that have well-defined values
over all states of such problems as:
\begin{alignat}{1}
  \label{eq:generalV}
  V(s)\ =\ F(\phi_1(s), \ldots, \phi_k(s)) \,.
\end{alignat}

\medskip
\noindent Linear value functions have the form
\begin{alignat}{1}
  \label{eq:linearV}
  V(s)\  =\ \sum\nolimits_{1\leq i\leq k}  w_i \phi_i(s)
\end{alignat}
where the coefficients $w_i$ are constants that do not depend on the states.
For example, a general, linear  value function for clearing block $x$ while having an empty gripper
is $V = 2 n + H$, where the states  are left implicit, and the Boolean feature $H$ is assumed to have value $1$ when true, and
$0$ otherwise.

Linear value functions using description logic features \citep{bonet:aaai2019},
called generalized potential heuristics, can be learned from small instances via a mixed integer programming formulation,
leading to an alternative representation of general policies that solve many standard
planning domains \citep{frances:ijcai2019}.

\section{Features}

Logical features derived from the domain predicates using a description logic grammar have  been used to
define and learn \emph{policies} of the form (\ref{eq:generalP}) and   \emph{value functions} of the form (\ref{eq:linearV}).\footnote{%
These logical features have also been used to encode ``sketches'', a generalization of policies that split
problems into (polynomial) subproblems of bounded width \citep{drexler:kr2021}.
Policies are a special type of sketches where the subproblems can be solved in one step \citep{bonet:aaai2021}.}
The complexity of such features is defined in terms of the number of grammar rules required to derive them,
and the pool of features used is obtained by placing a bound on the complexity of the features.
An important  limitation of these methods is that the pool of features grows exponentially with the complexity bound,
and that some domains  require complex features. For example, \citet*{frances:ijcai2019}
cannot learn general value functions for Logistics and Blocks because
they appear to require features of complexity 22 and 49, respectively. Interestingly, the features required to express the policy
rules for some of these domains is much smaller \citep{frances:aaai2021}.

For  learning general  policies without using a precomputed pool of features,
it  turns out to be  simpler and more direct to learn  general value functions,
and then define greedy policies from them.  A first step in this direction
was taken by \citet*{simon:arxiv2021} where the value function  $V$
was learned in a supervised fashion using graph neural networks from optimal targets $V^*$. Graph neural networks
have also been used in other approaches to generalized planning using deep nets \citep{sylvie:asnet,mausam:dl,karpas:generalized},
but in combination with other techniques  and without drawing on the relation between the features
that can be learned by GNNs and those that are actually  needed.
\section{Graph Neural Networks}

The GNN architecture for learning value functions follows the one used by \citet*{simon:arxiv2021}: it accepts
states $s$ over arbitrary instances of a given planning domain, and outputs
the scalar value  $V(s)$. For this, the form of the general value function $V(s)$
in \eqref{eq:generalV} is reformulated as:%
\begin{alignat}{1}
  \label{eq:embeddingV}
  V(s)\ =\ F(\phi(o_1), \ldots, \phi(o_n))
\end{alignat}
where $o_1, \ldots, o_n$ represent the   objects in the instance where the
state $s$ is drawn from,  $\phi(o)$ is a vector of feature values associated with object $o$ in state $s$
(dependence on $s$ omitted), represented as a vector of real numbers, and $F$ is a  function that aggregates these feature vectors
and   produces the scalar output $V(s)$. The vectors $\phi(o)$ are usually called \emph{object embeddings}
and the function $F$, the readout. Before revising the details of the architecture, it is worth discussing
the meaning and the implication of the transition from the fully  general value function form expressed  in \eqref{eq:generalV}
to the specific form expressed in  \eqref{eq:embeddingV}.

\subsection{From State Features to Object Embeddings}

We are moving from state features to object features $\phi(o)$ that depend not just on the state $s$ but  on the objects $o$.
In addition,  the same feature function $\phi$ is applied to all the objects, and
the same aggregation function $F$ is applied  to the states $s$ of  any of the domain instances
so that the number of feature vectors $\phi(o)$  expands or contracts  according to the number of objects  in the instance.
This is key for having a well-defined  value function  over the whole  collection of domain  instances that
involve a  different numbers of objects, not necessarily bounded.

The reasons for why  the restricted value function form \eqref{eq:embeddingV} is
rich  enough for capturing the value functions needed for generalized planning can be understood
by comparing \eqref{eq:embeddingV} with  the linear value functions \eqref{eq:linearV}
used by \citet*{frances:ijcai2019} in combination with description  logic features. These Boolean and numerical features
$b_q(s)$ and $n_q(s)$ are defined in terms of derived unary predicates $q$, where $b_q(s)=1$ (true) if there is an object $o$ such that $q(o)$ is true in s, otherwise $0$;
and $n_q(s)=m$ is the number of objects $o$ for which $q(o)$ is true in $s$. Clearly, if the feature vectors $\phi(o_i)$ in
\eqref{eq:embeddingV} contain a bit encoding whether $q(o)$ is true in $s$, then the readout function $F$ would just need to take the
\emph{max}  and the \emph{sum} of the bits $q(o)$ as
\begin{alignat}{1}
  b_q(s)\ &=\  \max_{o}\ q(o) \,, \\
  n_q(s)\ &=\ \sum_o\,\ q(o)\,,
\end{alignat}
in order to capture such features, where the objects $o$ range over all the objects $o$ in the instance.
In other words, the object-embedding form \eqref{eq:embeddingV} is no less expressive than the linear form that uses
description logic features, provided that the feature vectors $\phi(o)$ are expressive enough to represent the bits $q_i(o)$
for unary predicates $q_i$ derived from the domain predicates using the description logic grammar.
This in  turn is known to be within the capabilities of standard, message passing GNNs, that can
capture  the properties that can be expressed in the guarded fragment of the variable logic with counting $\C_2$,
which includes the standard description logics \citep{barcelo:gnn}.

Below we follow the terminology of graph neural networks and  refer to graphs and not states, and
to  vertex embeddings $f(v)$ and not  object embeddings $\phi(o)$.
After  considering    standard  GNNs for undirected graphs,
we introduce  the generalization needed for dealing with the relational structures
represented by  planning states.

\subsection{GNNs on Graphs}

GNNs represent trainable, parametric, and generalizable  functions over graphs
\citep{gori:gnn,book:gnn} specified
by means of aggregate and combination  functions $\agg_i$
and $\comb_i$, and a  readout function $F$. For each vertex $v$ of the  input graph $G$, the
GNN maintains a state (vector) $f_i(v) \in \reals^k$, the vertex embedding,
$i=0, \ldots, L$, where $L$ is the number of iterations or layers.
The vertex embeddings $f_0(v)$ are fixed and
the embeddings $f_{i+1}$ for all $v$ are computed from the $f_i$  embeddings as:%
\begin{alignat}{1}
  \label{eq:msg-passing}
  f_{i+1}(v) := \comb_i\bigl(f_i(v), \agg_i\bigl( \multiset{f_i(w) | w{\in}N_G(v) } \bigr) \bigr)
\end{alignat}
where $N_G(v)$ is the set of neighbors for vertex $v$ in $G$, and
$\multiset{\ldots}$ denotes a multiset.
In words, the embeddings  $f_{i+1}(v)$ at  iteration $i+1$ are  obtained by combining the
aggregation of  neighbors' embeddings $f_i(w)$ at iteration $i$
with $v$'s own embeddings $f_i(v)$. This process is usually seen
as an exchange of messages among neighbor nodes in the graph.
The aggregation functions  $\agg_i$ map arbitrary collections of
real vectors of dimension $k$ into a single $\reals^k$ vector.
Common aggregation functions are \emph{sum}, \emph{max}, and
\emph{smooth-max} (a smooth approximation of the max function).
The combination functions $\comb_i$ map pairs of $\reals^k$ vectors
into a single $\reals^k$ vector.  The  embeddings $f_{L}(v)$
in the last layer are  aggregated and mapped into the output of the GNN by
means of a readout function $F$. In our setting, the output will be a scalar $V$,
and the  aggregation and combination functions $\agg_i$ and $\comb_i$ will be homogeneous
and not depend on the layer index $i$. All the functions are
parametrized with weights that are adjusted by minimizing a suitable loss function.
By design, the function computed by a GNN is \emph{invariant} with respect to graph isomorphisms,
and once a  GNN is trained, its output is well defined for any graph $G$ regardless size.

\subsection{GNNs for Planning States}

\begin{algorithm}[t]\footnotesize
  \SetAlgoLined
  \KwIn{State  $s$: set of atoms true in $s$, set of objects}
  \KwOut{V(s)}
  \BlankLine
  $f_0(o)  \sim \mathbf{0}^{k/2}\mathcal{N}(0, 1)^{k/2}$ for each object $o \in s$\; \label{gnn:init}
  \For{$i \in \{0, \dots, L-1\}$}
  {
    \For{each atom $q := p(o_1, \dots, o_m)$ true in $s$}
    {
      \tcp{\footnotesize Msgs $q \rightarrow o$ for each $o=o_j$ in $q$}
              ${m}_{q,o}  := [\mlp_p(f_i(o_1), \ldots, f_i(o_m))]_j$\;\label{gnn:message}
    }
    \For{each $o$ in $s$}
    {
      \tcp{\footnotesize Aggregate, update embeddings}
                   $f_{i+1}(o) \!:=\!  \mlp_U\bigl(f_i(o), \agg(\multiset{{m}_{q,o} | o\in q })\bigr)$\; \label{gnn:update}
    }
  }
  \tcp{\footnotesize Final Readout} \label{gnn:readout}
    ${V} := \mlp_2\bigl(\sum_{o \in s}\mlp_1(f_L(o))\bigr)$
  \caption{GNN  maps  state $s$ into scalar $V(s)$}
  \label{alg:architecture}
\end{algorithm}

States $s$ in planning do not represent graphs but more general relational structures
that are defined by the set objects, the set of domain predicates, and the atoms
$p(o_1,\ldots,o_m)$ that are true in the state: the objects define the universe,
the domain predicates, the relations, and the atoms,  their denotations.
The  set of predicate  symbols $p$ and their  arities are fixed
by the  domain, but the sets of objects $o_i$ may change from
instance to instance. The adaptation of the basic GNN architecture for dealing with planning states $s$
follows \citep{simon:arxiv2021}, which is an elaboration of the architecture for
learning to solve Max-CSP problems over a fixed class of binary relations introduced by \citet*{grohe:max-csp}.
The new GNN still maintains just the object embeddings $f_i(o)$  for each of the objects $o$ in the input
state $s$, $i=0, \ldots, L$, but now rather than messages flowing from ``neighbor'' objects to objects
as in \eqref{eq:msg-passing},  the messages flow from objects $o_i$
to the true atoms $q$ in $s$ that include $o_i$,  $q=p(o_1, \ldots, o_m)$, $1  \leq i \leq m$,
and from such atoms $q$ to all the objects $o_j$  involved in $q$ as:
\begin{alignat}{1}
  \label{eq:msg-passing:2}
  f_{i+1}(o) := \comb_U \bigl(f_i(o), \agg\bigl( \multiset{m_{q,o} | o\in q, q \in s} \bigr) \bigr)
\end{alignat}
where $m_{q,o}$ for $q=p(o_1,\ldots,o_m)$ and $o=o_j$ is:
\begin{alignat}{1}
  \label{eq:msg-atoms}
  {m}_{q,o}\ :=\ [\comb_p (f_i(o_1), \ldots, f_i(o_m))]_j \,.
\end{alignat}
In these updates, the combination function  $\comb_U$ takes the concatenation of two real vectors of size $k$
and outputs a vector of size $k$, while the combination function $\comb_p$,  that depends on the predicate symbol $p$,
takes the concatenation of $m$ vectors of size $k$, where $m$ is the arity of $p$,
and outputs $m$ vectors of size $k$ as well, one for each object  involved in the $p$-atom.
The expression $[\ldots]_j$ in \eqref{eq:msg-atoms} selects the $j$-th such vector in the output.

The resulting trainable function that maps states $s$ into their values $V(s)$ is
shown in Algorithm~\ref{alg:architecture} with all the combination functions
replaced by the  multilayer perceptrons (MLPs) that implement them.
During the iterations $i=0, \ldots, L$,  a single $\mlp_U$ is used for updating the
object embeddings following \eqref{eq:msg-passing}, and a single $\mlp_p$ per predicate
is used to collect the messages from atoms to objects as in \eqref{eq:msg-atoms}.
The readout function, the last line in Algorithm~\ref{alg:architecture}, uses two MLPs
and a sum aggregator.
Finally, for the aggregator in line 6, we use the differentiable smooth max function
$smax(x_1, \dots, x_n)$ defined as
\begin{alignat}{1}
  x^* + \alpha^{-1} \log \left( \sum\nolimits_{1\leq j\leq n} \exp(\alpha (x_j - x^*)) \right)
\end{alignat}
where $x^* = \max\{x_1, \dots, x_n\}$ and $\alpha = 8$.

All MLPs consists of a dense layer with a ReLU activation function,
followed by a dense layer with a linear activation function.
The hyperparameter in the networks are the embedding dimension $k$
and the number of layers $L$. The initial embeddings $f_0(o)$
are  obtained by concatenating a  zero vector with a random
vector, each of dimension $k/2$, to break symmetries.
Random initialization increase expressive power for instances of fixed size~\citep{grohe:expressive}, however, we aim to learn policies for arbitrary sizes.
Key for the GNN to apply to any state over the domain is the use of
a single MLP$_p$ for each predicate symbol $p$ in the domain.

\section{Learning the GNN Parameters}

The parameters of the network displayed in Algorithm~\ref{alg:architecture}
are learned by stochastic gradient descent by minimizing a loss function.
In the work of \citet*{simon:arxiv2021}, the training data ${\cal D}$ is a collection of pairs $\tup{s,V^*(s)}$
for sampled states $s$ from selected instances,
and $V^*(s)$ is  the optimal cost for   reaching  the goal from $s$  (min.\ number of steps). The loss is  the
average sum of the  differences
\begin{equation}
  L(s)\ =\ |V(s)-V^*(s)|
\end{equation}
over the states $s$ in the training set. The computation of the optimal targets
$V^*(s)$ is not a problem because we are  computing them over small instances.
The real   problem is that by  forcing  the  value function to be optimal over the training instances,
domains such as Blocks or Miconic, where optimal planning  is  NP-hard \citep{blocks-np-hard,miconic-np-hard},
are excluded (except when the  goals are restricted to be  single atoms).

Interestingly, as discussed in the next section,  this limitation  pops up also in unsupervised,
reinforcement learning  approaches where the optimal target values $V^*(s)$ are not given
but are sought by minimizing the Bellman error:
\begin{equation}
  L'_0(s)\ =\ |V(s) - (1 + \textstyle\min_{s' \in N(s)} V(s'))|
\end{equation}
for non-goal states $s$, where $N(s)$ are the states reachable from $s$ in one step (possible successor states).
For goal states, $L'_0(s)$ is $|V(s)|$. The optimal function $V^*$ is  the unique value function
that minimizes the resulting loss, provided that actions costs are all $1$ and the goal is reachable from
all states. In this work, rather than penalizing departures from the Bellman optimality equation
\begin{equation}
  V(s)\ =\ 1 + \textstyle \min_{s' \in N(s)} V(s')\,,
\end{equation}
departures from the inequality $V(s) \ge 1 + \min_{s'\in N(s)} V(s')$ are penalized
with a loss for non-goal states $s$ defined as
\begin{equation}
  L'_1(s)\ =\ \max\{0, (1 + \textstyle \min_{s' \in N(s)} V(s')) - V(s)\} \,.
\end{equation}
Furthermore, this loss is extended with two regularization terms that penalize large departures from
$V^*$; namely, as done by \citet*{frances:aaai2021}, we want a value function $V$ that also satisfies
$V^* \leq V \leq \delta V^*$, and thus settle for the minimization of the loss:
\begin{alignat}{1}
  \notag
  L_1(s)\ =\  L'_1(s)\ +\ &\max\{0, V^*(s)-V(s)\} \ + \\
                          &\max\{0, V(s)-\delta V^*(s)\} \,,
  \label{eq:l1}
\end{alignat}
where $\delta=2$.
The loss over a set $\cal S$ of states is the sum of the average of $L_1(s)$ for non-goal states $s \in \cal S$ and the average of $|V(s)|$ for goal states $s \in \cal S$.
For comparison purposes, the $L'_0$ loss is extend into
the regularized $L_0$ loss as well as:
\begin{alignat}{1}
  \notag
  L_0(s)\ =\  L'_0(s)\ +\ &\max\{0, V^*(s)-V(s)\} \ + \\
                          &\max\{0, V(s)-\delta V^*(s)\} \,,
  \label{eq:l0}
\end{alignat}

If all the states in a small instance are in $\cal S$ and the overall loss is close to zero,
the loss function $L_1$ results in value functions that lead greedily to the goal
(by picking the min-$V$ successors), while the loss $L_0$ results in value functions that lead greedily
and \emph{optimally}  to the goal.  For simplicity, it is assumed   that the domains considered
do not have \emph{dead-ends}, i.e.\ states from which the goal is not reachable and where  $V^*(s)$ is not well-defined.
Learning to plan in such domains requires an slight extension, with extra  inputs, for labeling states as  dead-ends in the training data,
and  extra outputs, for predicting if a state is a dead-end \citep{simon:ijcai2021}. This extension is  implemented and tested, but it will be skipped
over in the presentation.

\section{Experiments}

The experiments are aimed to test the generalization,  coverage, and quality of the plans
obtained by the policy $\pi_V$ greedy in the learned value function $V$, using the unsupervised losses $L_0$
and $L_1$. We describe the training and testing data used, and  the results.
A key difference with prior work \citep{simon:arxiv2021} is that the test instances
are standard IPC planning problems from standard planning domains, several of which
are intractable for optimal planning. We seek crisp experimental results, which
means close to $100\%$ generalization,  or alternatively, crisp explanations of why
this is not possible, with  logical fixes that restore  generalization   in certain
cases.

\subsubsection*{Data.}
The states  in the training and validation sets  are obtaining by  fully expanding
selected instances from the initial state through a breadth-first search.
For each  reachable state, the length of the shortest path to a goal state is
computed. For instances with large state spaces we  keep up to $40,000$ sampled reachable states
to avoid large instances from dominating the training set. The actual size of the instances
used in training, validation, and testing are shown in Table~\ref{tbl:experiments:sizes},
measured by the number of objects involved. In almost all cases, the testing instances
are IPC (International Planning Competition) instances. The exception is the domain Spanner*,
which is a slight variant of the Spanner domain that does not give rise to dead-end states
by allowing the agent to move not just forward but also  backward.

\begin{table}[t]
  \centering
  \small
  \begin{tabular}{@{}lccc@{}}
  \toprule
  Domain    & Train        & Validation     & Test           \\ \midrule
  Blocks    & {[}4, 7{]}   & {[}8, 8{]}     & {[}9, 17{]}    \\
  Delivery  & {[}12, 20{]} & {[}28, 28{]}   & {[}29, 85{]}   \\
  Gripper   & {[}8, 12{]}  & {[}14, 14{]}   & {[}16, 46{]}   \\
  Logistics & {[}5, 18{]}  & {[}13, 16{]}   & {[}15, 37{]}   \\
  Miconic   & {[}3, 18{]}  & {[}18, 18{]}   & {[}21, 90{]}   \\
  Reward    & {[}9, 100{]} & {[}100, 100{]} & {[}225, 625{]} \\
  Spanner*  & {[}6, 33{]}  & {[}27, 30{]}   & {[}22, 320{]}  \\
  Visitall  & {[}4, 16{]}  & {[}16, 16{]}   & {[}25, 121{]}  \\ \bottomrule
  \end{tabular}
  \caption{Instance sizes used training, validation, and testing datasets, as
    measured by the number  of objects involved. E.g., the training set for
    Blocks consists of IPC instances with a number of blocks between 4 and 7.
    There is no instance that is in more than 1 set (same number of objects, initial state and goal description).
  }
  \label{tbl:experiments:sizes}
\end{table}

\subsubsection*{Domains.}
The domains are those used by  \citet*{frances:aaai2021} with the addition of Logistics,
and the above modification of Spanner. Briefly, Blocks is the standard blocks
world.  Delivery is the problem of picking up objects in an empty grid and delivering
them one by one to a target cell.  Gripper is about moving balls from one room to another
with a moving robot that can have more than one gripper. Logistics involves
trucks and airplanes that move within city locations  and across cities,
where packages have to be moved from one location to another location, possibly in a different city.
Miconic is about controlling an elevator to pick up passengers in different floors
to their destination floors. Rewards is about reaching certain cells in a grid while avoiding
others. Spanner is about collecting spanners spread in a one dimensional grid,
each one to be  used to tighten up a single nut at the other end. Visitall
is about visiting  all  or some cells in an empty grid.

\subsubsection*{Setup.}
The hyperparameters $k$ and $L$ in Algorithm~\ref{alg:architecture} are set to $64$ and $30$,
respectively: $k$ is the number of ``features''  per object; i.e., the size of the real object embedding
vectors; and $L$ the number of layers in the GNN (fixed for training and testing). Both hyperparameters
affect training speed,  memory, and  generalization. Hyperparameter $L$ affects how far messages can propagate in the
graph, and indeed,  the GNN  cannot capture   shortest paths  between two objects if longer  than $L$, even
if the existence of paths up to length $2L$ can be determined. The architecture is implemented in PyTorch~\citep{pytorch} and
the optimizer  Adam~\citep{kingma15} is used
with a learning rate of $0.0002$.\footnote{Code and data: \url{https://doi.org/10.5281/zenodo.6511809}}
The networks are trained with NVIDIA A100 GPUs for up to 12 hours.
Five  models for each domain are trained  to ensure that the optimizer did not get stuck
in ``bad'' local minima, and the final model used is  the one with the best validation loss
(i.e., loss measured on the validation set). The quality of the plans obtained by following the greedy policy $\pi_V$ for the learned value function $V$
are evaluated in comparison with  optimal plans that are computed  with the  Fast Downward (FD) planner \citep{helmert:fd} using the
\emph{seq-opt-merge-and-shrink} configuration with time  and memory outs  set to $10$ minutes and $64$ GB, respectively,
on a Ryzen 9 5900X CPU.

\begin{table*}[t]
  \centering
  \small
  \begin{tabular}{@{\extracolsep{-1pt}}llrllrl@{}}
    \toprule
                                     & \multicolumn{3}{c}{\bf Deterministic policy $\pi_V$ with cycle avoidance}   & \multicolumn{3}{c}{\bf Deterministic policy $\pi_V$ alone} \\
                                 \cmidrule(lr){2-4} \cmidrule(lr){5-7}
    Domain (\#)                  & Coverage (\%) & L     & PQ = PL / OL (\#)         & Coverage (\%) & L     & PQ = PL / OL (\#)         \\
    \midrule \\
    [-.5em]
    \multicolumn{7}{c}{\bf $L_1$ Loss} \\
    \midrule \midrule
                                             Blocks  (20) &  20 (100\%) &    790 &  1.0427 =    440 /    422  (13) &  20 (100\%) &    790 &  1.0427 =    440 /    422  (13) \\
                                       Delivery  (15) &  15 (100\%) &    400 &  1.0000 =    400 /    400  (15) &  15 (100\%) &    404 &  1.0100 =    404 /    400  (15) \\
                                        Gripper  (16) &  16 (100\%) &  1,286 &  1.0000 =    176 /    176   (4) &  16 (100\%) &  1,286 &  1.0000 =    176 /    176   (4) \\
                                      Logistics  (28) &  17  (60\%) &  4,635 &  9.7215 =  3,665 /    377  (15) &   0   (0\%) &      0 &                             --- \\
                                        Miconic (120) & 120 (100\%) &  7,331 &  1.0052 =  1,170 /  1,164  (35) & 120 (100\%) &  7,331 &  1.0052 =  1,170 /  1,164  (35) \\
                                         Reward  (15) &  11  (73\%) &  1,243 &  1.2306 =  1,062 /    863  (10) &   3  (20\%) &    237 &  1.1232 =    237 /    211   (3) \\
                                    Spanner*-30  (41) &  30  (73\%) &  1,545 &  1.0000 =  1,545 /  1,545  (30) &  24  (58\%) &    940 &  1.0000 =    940 /    940  (24) \\
                                       Visitall  (14) &  14 (100\%) &    904 &  1.0183 =    556 /    546  (10) &  11  (78\%) &    631 &  1.0107 =    471 /    466   (9) \\
\midrule
                                          Total (269) & 243  (90\%) & 18,134 &  1.6410 =  9,014 /  5,493 (132) & 209  (77\%) & 11,619 &  1.0156 =  3,838 /  3,779 (103) \\
%
 \\
    [-.5em]
    \multicolumn{7}{c}{\bf $L_0$ Loss} \\
    \midrule \midrule
                       Blocks  (20) &   0   (0\%) &      0 &                             --- &   0   (0\%) &      0 &                             --- \\
                 Delivery  (15) &  12  (80\%) &    278 &  1.0000 =    278 /    278  (12) &  12  (80\%) &    278 &  1.0000 =    278 /    278  (12) \\
                  Gripper  (16) &  16 (100\%) &  1,288 &  1.0000 =    176 /    176   (4) &  12  (75\%) &    816 &  1.0000 =    176 /    176   (4) \\
                Logistics  (28) &   1   (3\%) &    134 & 16.7500 =    134 /      8   (1) &   0   (0\%) &      0 &                             --- \\
                  Miconic (120) & 120 (100\%) &  7,758 &  1.0241 =  1,192 /  1,164  (35) & 108  (90\%) &  6,438 &  1.0000 =  1,084 /  1,084  (33) \\
                   Reward  (15) &  12  (80\%) &  1,362 &  1.1226 =    861 /    767   (9) &   7  (46\%) &    661 &  1.0285 =    505 /    491   (6) \\
              Spanner*-30  (41) &  24  (58\%) &  1,221 &  1.0374 =  1,221 /  1,177  (24) &  14  (34\%) &    475 &  1.0000 =    475 /    475  (14) \\
                 Visitall  (14) &  14 (100\%) &    838 &  1.0073 =    550 /    546  (10) &  12  (85\%) &    664 &  1.0073 =    550 /    546  (10) \\
\midrule        
                    Total (269) & 199  (73\%) & 12,879 &  1.0719 =  4,412 /  4,116  (95) & 165  (61\%) &  9,332 &  1.0059 =  3,068 /  3,050  (79) \\

 \\
    [-.5em]
    \multicolumn{7}{c}{\bf  Derived Atoms ($L_1$ Loss)} \\
    \midrule \midrule
                  Logistics-atoms   (28) &  28 (100\%) &  8,147 &  5.5711 =  2,546 /    457  (17) &   4  (14\%) &     88 &  1.0353 =     88 /     85   (4) \\
              Spanner*-10       (36) &  12  (33\%) &    557 &  1.0000 =    557 /    557  (12) &   8  (22\%) &    373 &  1.0000 =    373 /    373   (8) \\
              Spanner*-atoms-5  (36) &  31  (86\%) &  1,370 &  1.0000 =  1,112 /  1,112  (27) &  28  (77\%) &  1,190 &  1.0000 =    996 /    996  (25) \\
              Spanner*-atoms-2  (36) &  36 (100\%) &  1,606 &  1.0000 =  1,348 /  1,348  (32) &  36 (100\%) &  1,606 &  1.0000 =  1,348 /  1,348  (32) \\
\midrule
                         Total (136) & 107  (78\%) & 11,680 &  1.6013 =  5,563 /  3,474  (88) &  76  (55\%) &  3,257 &  1.0011 =  2,805 /  2,802  (69) \\[1em]
%

    \bottomrule
  \end{tabular}
  \caption{Performance of the deterministic greedy policy $\pi_V$  for the learned value function $V$ when executed with cycle avoidance (left) and without (right).
    Three subtables shown: results when using the $L_1$ loss (top), results when using the $L_0$ loss (middle), and results using $L_1$ loss when states are extended with derived     atoms (encoding role compositions and transitive closures). The domains are  shown on the left with the number of instances tested in each.
    Coverage is the number of solved problems.
      L is the sum of the solution lengths over the test instances solved by the learned policy.
      PQ is a measure of  overall plan quality  given by the ratio of the sum of the plan lengths found by the policy  (PL) and the  optimal plan lengths (OL)  found by FD,
      over the  instances solved by both within the time and memory limits (number of problems solved by FD   shown after OL in parenthesis).
  }
  \label{tbl:experiments:coverage}
\end{table*}

\subsection{Testing the Greedy Policy $\pi_V$: Two Modes}

The  greedy policy $\pi_V$ selects the action applicable in a non-goal state  $s$
that leads to  the child state $s'$ with minimum $V(s')$  value (action costs are all assumed to be $1$).
It is common to add ``noise'' in this selection process by either
breaking ties randomly or by choosing the action leading to the best
child probabilistically, by soft-mapping the children values $V(s')$
into probabilities that add up to $1$. The addition of ``noise''
in action selection has the benefit that it helps to avoid
cycles in the execution, but at the same time,
it blurs the results. Instead, Table ~\ref{tbl:experiments:coverage}  shows (on the right)  the
results of the executions that follow the deterministic greedy policy $\pi_V$,
which  always chooses the action leading to the child $s'$ with lowest $V(s')$ value,
breaking ties for the first such action encountered. Since the learned value function
is not perfect, we show on the left the execution of the greedy policy but with
\emph{cycle avoidance}; namely,  executions keep track of the visited states
and deterministically select  the first action leading to the  best \emph{unvisited} child (min-$V$ value).
When there are no such children, the execution fails. Executions are also terminated
when the goal is not reached  within $1,000$ steps.

\subsection{Results: $L_1$ Loss}

Table~\ref{tbl:experiments:coverage} shows the results for various experiments:
learning using  the $L_1$ loss (top), learning using the $L_0$ loss (middle),
and learning using states augmented with derived atoms  in domains that
benefit from \CCC features  (explained below).
Furthermore, the three subtables are divided horizontally in two,
according to the way in which the greedy policy $\pi_V$ for the learned value function $V$ is
used: with cycle avoidance, on the left, and without cycle avoidance, on the right.
We focus now on the top part of the table.

\subsubsection*{Coverage.}
The first thing to notice is that in 4 out of the 8 domains considered,
Blocks, Delivery, Gripper, and Miconic, the deterministic greedy policy $\pi_V$ for the learned value function
$V$ solves all the test instances. This is pretty remarkable as the resulting plans are often long.
In Blocks, the average plan length is $790/20=39.5$ steps, while in Miconic, it is $7,331/120=61.09$.
As we will see, while the plans are not optimal, they are very good, and moreover, in none of these
cases, the deterministic greedy policy generates an execution where a state is revisited.
Indeed, if revisits are explicitly excluded by executing the greedy policy while avoiding cycles (left),
a fifth domain is solved in full: Visitall. The other three domains  are not solved in full in either
mode: Logistics, Reward, and Spanner. In the case of Logistics, the reason, as we will see,
is purely logical: given the domain representation of Logistics, the feature expressing
that a package is in a location or in a city, while possibly within a vehicle,
involves the composition of two or three  binary relations, requiring  three variables,
which  is not possible in   \CC. We address this expressive limitation of GNNs below by adding suitable
``derived'' atoms to the state
that bypass the need for such compositions. The limitations observed in Reward and Spanner
are not logical: these two domains, as others in the list, require the computation of distances
to determine in which direction to move (e.g., to the nearest reward or right exit).
Yet GNNs  cannot compute distances that exceed their  number of layers $L$.
Actually, there are other domains solved in full that require the computation of distances,
but the magnitude of the  distances needed in the test set does not
defy these bounds.  Indeed, even a simple problem such a clearing a block $x$ may be found
to be unsolvable by the learned policy if the number of blocks above $x$ is much larger
than $L$. Interestingly, this limitation  has  an easy  logical ``fix'' in some of the domains,
where derived atoms capturing the \emph{transitive closure}  of some binary predicates manage to decouple the computation of distances
from the number of layers in the GNN. In the domains where these expressive limitations arise, the greedy policy with cycle
avoidance does better than the  pure greedy policy, as the latter is more likely to be trapped in cycles.

\subsubsection*{Quality.}
Somewhat surprisingly, the quality of the executions delivered by the
models trained with the $L_1$ loss  is very  close to optimal, as measured with respect to
the optimal plans computed by FD. The only exception is  the  Logistics domain  where
plans are up to 10 times longer than  optimal,  on  average. These results are  surprising not
just because the $L_1$ loss does not force the  value function $V$ to be optimal, but
because  optimal planning in several of these domains, certainly Blocks,   Miconic, and Logistics, and possibly in Reward and Visitall as well,
is NP-hard \citep{blocks-np-hard,miconic-np-hard}.
For example, FD with the given time and memory bounds computes optimal solutions
for 35 instances in Miconic comprising a total of 1,164 actions, while the sum
of execution lengths for the learned, greedy policy $\pi_V$ with or without
cycle avoidance  on the same 35 instances is 1,170. Indeed, the  execution
lengths that follow from the learned value function do not exceed the optimal plan
lengths in more than $12\%$ with the exception of Logistics.

\subsection{$L_0$ Loss: General Policies and RL}

The differences between the $L_1$ loss (\ref{eq:l1})
and  the $L_0$ loss (\ref{eq:l0})  are small but significant.
Zero loss for $L_0$  arises just when the learned $V$ function
has zero Bellman error over the training set; i.e.\ when $V(s)=1+\min_{s'\in N(s)} V(s')$ for the possible children $s'$ of $s$,
and thus when $V$ is the optimal cost function $V^*$.
Zero loss for $L_1$, on the other hand, arises just  when the learned $V$ function is such that $V(s) \ge 1+\min_{s'\in N(s)} V(s')$.
Thus, zero $L_0$ loss implies zero $L_1$ loss, but not the other way around, as the $L_1$ loss captures
just one half of Bellman's  optimality equation.
Provided that only the goal states have zero value and that non-goal states have
positive values, one can use a value function $V$ with zero $L_1$ loss
to solve problems \emph{greedily} by always moving to the best child
(min $V$). On the other hand, a value function $V$ with zero $L_0$ loss
can be  used in the same manner to solve problems greedily and \emph{optimally}.
The difference between solving a class of problems optimally or suboptimally
is crucial in  domains where optimal planning is NP-hard. Such domains, like Blocks,
often  admit  general policies but no general  policies that are optimal.

So the question arises as to whether the minimization of the $L_0$ loss
leads to greedy policies $\pi_V$ that are as good as,  or better than
those obtained by minimization of the  $L_1$ loss. The question is particularly
relevant because the  standard methods for learning policies without
supervision are usually  based on reinforcement learning,  which
in their value-based variant (as opposed to the policy gradient version)
are based on the  minimization of Bellman error \citep{sutton:book}.
The expectation is that the minimization of $L_0$ loss will not be as good.
Indeed, the value functions $V$ that yield greedy  policies $\pi_V$
that generalize correctly over domains that are intractable for optimal planning
are unlikely to yield zero $L_0$ loss.

The middle  part of Table~\ref{tbl:experiments:coverage} shows the results
of  the greedy  policies $\pi_V$ for value functions $V$ learned by minimizing
$L_0$ loss instead of $L_1$. The $L_0$-based policies are observed to
perform worse  than the $L_1$-based policies. The extreme case is
precisely in Blocks where coverage drops from 100\% to 0\%
when using the greedy policy with cycle avoidance and also without.
A big difference also surfaces in Logistics where coverage
drops from 60\% to 3\% with cycle avoidance (otherwise
no instances are solved). For the other domains, the drops are not as drastic,  yet
the greedy policy with no cycle avoidance based on $L_1$  solves four domains fully
($100\%$ coverage) while the same policy based on $L_0$ does not solve fully
any single domain. The $L_0$-policies,  however,  do  slightly better in two of the domains
where the $L_1$-policy is not good: Reward and Visitall where coverage increases
from 20\% and 78\% to 46\% and 86\%. As expected, the lower coverage
of $L_0$-policies goes along with executions whose lengths  are better overall.
With cycle avoidance, the performance resulting from the two loss functions
is closer,  with the aforementioned exceptions.
In general, the ability of the learned value functions $V$ to yield greedy  policies that generalize
can be predicted from the corresponding loss on the validation set. In both  Blocks and Logistics,
the validation loss after $L_1$ training is close to zero, but significantly higher than zero after $L_0$ training.
\subsection{Derived Atoms: Beyond \CC}

The failure   of the learned policies to generalize fully when using the $L_1$ loss function
in domains such as Logistics, Reward, and Spanner* can be traced to two limitations.
Logistics requires features that cannot be expressed in \CC and which therefore
are not captured by GNNs \citep{barcelo:gnn,grohe:gnn}.
Spanner*, like Reward and other domains, involves the computation of distances
in the test instances that exceed the number of layers used in the GNN. The bottom part of
Table~\ref{tbl:experiments:coverage} shows the results that are obtained
in Logistics and Spanner* when these limitations are addressed logically
by extending the states (in training, validation, and testing) with
suitable derived  atoms and predicates, a facility provided by  PDDL
\citep{axioms:pddl,haslum:pddl}. For example, one can extend the states
in Blocks with the derived predicate $above$ that corresponds
to the transitive closure of the domain predicate $on$, so that every
state $s$ contains additional atoms $above(x,y)$ when block $x$ is above block $y$ in $s$.

In Logistics, four derived predicates are added, following the four role
compositions used by \citet*{frances:ijcai2019} to obtain a general value function.
These role compositions go  beyond the expressive capabilities of \CC and GNNs.
In Logistics, there are binary predicates (roles) to express that a package or truck is at some location (`at'),
to express that a package is inside a truck or airplane (`in'), and to express that a location is
in a city (`in-city'). Additionally, as done in previous works, ``goal versions'' of these
predicates (indeed, of all predicates) denoted by `at@', `in@' and `in-city@' whose denotation
is provided by the goal descriptions are added to the domain.
The Logistics domain is extended with the following role compositions from \citet*{frances:ijcai2019}:
\begin{enumerate}[--]
  \item `$\text{at}\!\circ\!\text{in-city}$' and `$\text{at@}\!\circ\!\text{in-city}$'
    that tells the city where a package is located, either in the current or goal state,
  \item `$\text{in}\!\circ\!\text{at}$' that tells the location of a package that is inside a truck, and
  \item `$\text{in}\!\circ\!\text{at}\!\circ\!\text{in-city}$' that tells the city where a package that is inside
    a truck is located.
\end{enumerate}

In Spanner*, a single derived predicate
is added which is  the transitive closure of the `link' predicate.
Provided with the new $\text{link}^+$ predicate, the  required distances
in Spanner*  are not restricted by  the number of layers $L$ in the GNN
and can be computed in a single layer, as the distance to the exit
location equals the number of locations to the right of the current location $c$; i.e., $\text{dist2exit}=|\{ x\,|\,\text{link}^+(c,x)\}|$.

The results obtained by learning from states with these derived predicates
in Logistics and Spanner* are shown at the bottom of Table~\ref{tbl:experiments:coverage}.
In Logistics, the simple addition of the atoms makes the coverage jump from
from 0\% to 14\% for the greedy policy alone, and from 60\% to 100\%
for the greedy policy with cycle avoidance. For Spanner*, three rows
are shown: the first is for the domain without derived atoms
but with two modifications that preclude comparison with
the Spanner* results reported previously in the same table.
The first is that the test instances involving more than 100 locations
have been replaced by smaller instances with up to 45 and 50 locations.
The second is that the number of layers $L$ in the GNN are reduced from $30$ to $10$.
These modifications provide a more convenient baseline for
evaluating the impact of derived atoms: with $100$ locations, there
are $10,000=100^2$ extra derived atoms in the states,
that make training and testing much slower (this is a weakness
of adding derived atoms). It is because of these modifications,
and in particular from the reduction in the value of $L$ from $30$ to $10$, that the coverage
of the learned policies in the modified Spanner* setting is reduced to 33\% and 22\% percent
(first of the last three rows in the table). This number however increases to 86\% and 77\%
when the derived atoms are included, even if the number of GNN layers is reduced from $10$ to $5$
(second of the last three rows in table). Moreover, coverage increases further to 100\%
when the derived atoms are included and the number of GNN layers is reduced further to just $2$ (last row in the table).
This additional increase in coverage is likely due by reduced overfitting
as the number of layers $L$ is reduced from $5$ to $2$.

\section{Conclusions}

We have  considered  the problem of learning generalized policies for classical planning domains
from small instances represented in lifted STRIPS. Unlike previous work that makes uses
of a predefined pool of features based on description logic and combinatorial solvers,
we have followed the  GNN approach for learning general policies advanced by
\citet*{simon:arxiv2021} that exploits the relation between \CC features and
those that can be computed by GNNs. However, instead of learning optimal value
functions in a supervised manner, we learn non-optimal value functions without
supervision. For this, the change  is technically small, as it affects
the loss function and not the GNN architecture, but the consequences
are interesting as the new method can be applied to domains
that have general policies but no general policies that are optimal.
We have shown that  100\% generalization is achieved in many such domains,
and have discussed and  addressed two important additional  issues:
the  limitations of value-based RL methods for computing
general policies over domains where optimal planning is intractable,
and the limitations of GNNs for capturing general value functions
that require non-\CC features. We have addressed the first limitation
by using a novel loss function ($L_1$) different than the more natural
loss function $L_0$  associated with value-based RL methods,
and the second limitation,  by extending  planning states with derived atoms.
In the future, we would like to make the  point about the limitations
of RL methods for learning generalized plans, sharper, and to consider
the use of recent GNN architectures that compute features beyond \CC
\citep{gnn:beyond}. At the same time, we are interested in ``domesticating''
the use of deep learning engines in the context of planning and representation learning
for planning, so that they can be used as alternatives to ASP and Weighted Max-SAT solvers,
for avoiding  scalability issues and  for opening up new possibilities.
This requires understanding what can be computed  with them in a clean way  and how.
This work is  also a step in that direction.

\section*{Acknowledgments}

The code framework Tarski~\citep{tarski} was very useful during this research.
This research was partially supported by the European Research Council (ERC), Grant No.\ 885107, and by project TAILOR, Grant No.\ 952215, both funded by the EU Horizon research and innovation programme.
This work was partially supported by the Wallenberg AI, Autonomous Systems and Software Program (WASP) funded by the Knut and Alice Wallenberg Foundation.
The computations were enabled by the supercomputing resource Berzelius provided by National Supercomputer Centre at Link\"{o}ping University and the Knut and Alice Wallenberg foundation.

{
\bibliography{paper}
\bibliographystyle{abbrvnat}
}

\end{document}